%% file: aaai25.tex
\documentclass[letterpaper]{article} 
\usepackage{aaai25}  
\usepackage{times}  
\usepackage{helvet}  
\usepackage{courier}  
\usepackage[hyphens]{url}  
\usepackage{graphicx} 
\usepackage{natbib}  
\usepackage{caption} 
\usepackage{algorithm}
\usepackage{algorithmic}

\usepackage{newfloat}
\usepackage{listings}
\DeclareCaptionStyle{ruled}{labelfont=normalfont,labelsep=colon,strut=off} 
\floatstyle{ruled}
\newfloat{listing}{tb}{lst}{}
\floatname{listing}{Listing}
\usepackage{colortbl}
\usepackage{amsmath}
\usepackage{amssymb}
\usepackage{xcolor}
\usepackage{booktabs}
\usepackage{multirow}
\usepackage{pifont}
\usepackage{makecell}
\usepackage{threeparttable}
\usepackage{bbding}

\title{Text and Image Are Mutually Beneficial: \\ Enhancing
Training-Free Few-Shot Classification with CLIP}
\author{
    Yayuan Li\textsuperscript{\rm 1},
    Jintao Guo\textsuperscript{\rm 1},
    Lei Qi\textsuperscript{\rm 2},
    Wenbin Li\textsuperscript{\rm 1},
    Yinghuan Shi\textsuperscript{\rm 1}\thanks{Corresponding author.}
}
\affiliations{
    \textsuperscript{\rm 1}National Key Laboratory for Novel Software Technology, Nanjing University, China\\
    \textsuperscript{\rm 2}School of Computer Science and Engineering, Southeast University, China\\
    \{liyayuan, guojintao\}@smail.nju.edu.cn, qilei@seu.edu.cn, \{liwenbin, syh\}@smail.nju.edu.cn
}

\begin{document}

\maketitle

\begin{abstract}
Contrastive Language-Image Pretraining (CLIP) has been widely used in vision tasks. Notably, CLIP has demonstrated promising performance in few-shot learning (FSL).
However, existing CLIP-based methods in training-free FSL (\emph{i.e.,} without the requirement of additional training) mainly learn different modalities independently, leading to two essential issues: 1) severe anomalous match in image modality; 2) varying quality of generated text prompts.
To address these issues, we build a mutual guidance mechanism, that introduces an Image-Guided-Text (IGT) component to rectify varying quality of text prompts through image representations, and a Text-Guided-Image (TGI) component to mitigate the anomalous match of image modality through text representations.
By integrating IGT and TGI, we adopt a perspective of \textbf{T}ext-\textbf{I}mage \textbf{M}utual guidance \textbf{O}ptimization, proposing TIMO.
Extensive experiments show that TIMO significantly outperforms the state-of-the-art (SOTA) training-free method. Additionally, by exploring the extent of mutual guidance, we propose an enhanced variant, TIMO-S, which even surpasses the best training-required methods by 0.33\% with approximately \textbf{×100} less time cost. Our code is available at \textcolor{magenta}{https://github.com/lyymuwu/TIMO}.

\end{abstract}


\section{Introduction}

Given its strong zero-shot capability, Contrastive Language-Image Pretraining (CLIP) \citep{Radford_2021_Learning} has recently been adopted for various downstream tasks \citep{
Shao_2024_DeIL, Tang_2024_AMU-Tuning, Martin_2024_Transductive, Zhou_2023_ZegCLIP}. Among these, CLIP-based few-shot learning (FSL) has garnered significant attention due to its ability to quickly adapt to downstream tasks with few training samples \citep{Wang_2020_Generalizing, Sanghi_2022_CLIP-Forge, Zhao_2024_CLIP, Mu_2022_SLIP, Wang_2024_Connecting}.
However, as models and datasets grow, traditional training-required FSL methods face significant time costs due to backpropagation and are prone to overfitting, diminishing the rapid adaptability of FSL. Consequently, training-free FSL methods \citep{Zhang_2022_Tip-Adapter, Zhu_2023_Not, Wang_2024_Hard-to-Beat} have emerged, aiming to improve model performance while avoiding iterative optimization methods like gradient descent to enhance efficiency.

\begin{figure}[!ht]
    \centering
    \includegraphics[width=1\linewidth]{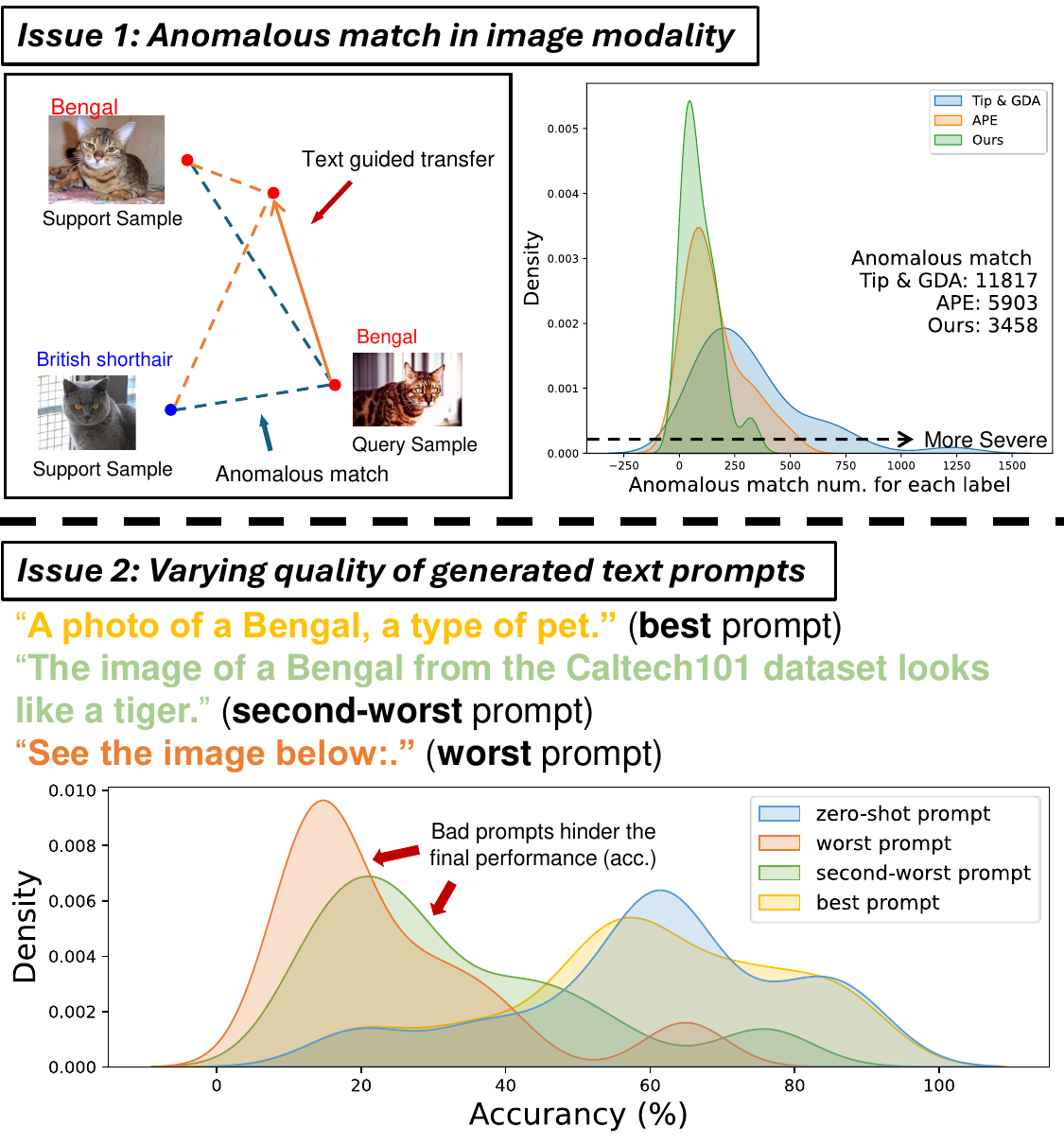}
    \caption{Defects of independent modelling. The top of the figure illustrates how the anomalous match left from the CLIP pretraining stage results in high similarity between images from distinct categories; The bottom shows the varying quality of the automatically generated prompts, where poor prompts can negatively impact the final performance.} 
    \label{Illustration}
\end{figure}

Despite the promising performance of previous state-of-the-art (SOTA) training-free FSL methods, they generally model different modalities separately, ignoring the importance of effective cross-modal interaction (\emph{e.g.,} modality alignment). This is particularly challenging to achieve in training-free settings, as we cannot update model parameters as in conventional methods. Consequently, previous related works either only focus on designing sophisticated architectures \citep{Wang_2024_Hard-to-Beat, Zhang_2022_Tip-Adapter, Udandarao_2023_SuS-X, Zhang_2023_Prompt}, or only concentrate designing prompts to make them more adaptable to image classification tasks \citep{Zhou_2022_Learning, Lu_2022_Prompt, prattWhatDoesPlatypus2023, Roth_2023_Waffling}.
This widespread lack of inter-modal interaction leads to previous SOTA models failing to capture the interrelation of representations from various sources, resulting in suboptimal classifiers \citep{Gong_2017_Exploring, Tan_2021_Individuality}. This observation motivates us to deeply investigate the consequences of inadequate cross-modal interaction and explore potential solutions with the following two bottlenecks.

\textbf{1) Severe anomalous match in image modality}. Prior works have highlighted that different modalities possess inherent strengths and weaknesses \citep{Fan_2023_PMR, Zhang_2024_Multimodal}. Notably, we found that the information contained in a single image is significantly more abundant than that in text, which makes the CLIP's inherent flaws (caused at the pretraining stage) of the image modality more pronounced. Besides, this issue is also particularly pronounced in FSL with scarce training samples, as it is often masked by the large training datasets in fully supervised settings.
For instance, limited support images tend to lead to high confidence in incorrect predictions, which is a specific manifestation of the inherent weakness. This issue was observed in our task, as shown in Fig. \ref{Illustration}, where a 1-shot classification mispredicts a \textit{Bengal} as a \textit{British Shorthair}. To investigate further, we conducted a statistical analysis of the \textit{OxfordPets} dataset. 
We define an anomalous match as a high degree of similarity between the prototype feature of one category and the image feature of another category.
Our findings reveal that anomalous matches are prevalent, as illustrated in the density plot at the top of Fig. \ref{Illustration}, where the prototype features of most categories have hundreds of anomalous matches. This issue is particularly severe for methods like Tip-Adapter \citep{Zhang_2022_Tip-Adapter} and GDA-CLIP \citep{Wang_2024_Hard-to-Beat}, which have not considered inter-modal alignment. In contrast, APE \citep{Zhu_2023_Not} mitigates this problem to some extent through dimensionality reduction. Ultimately, this phenomenon significantly increases the likelihood of misclassification in subsequent tasks. 
Here, we raise a question: \textit{Could the text be well utilized to mitigate the anomalous match in image modality?}

\textbf{2) Varying quality of generated prompts.} 
Previous training-free methods produce prompts by independently generating descriptive and structured prompts \citep{Roth_2023_Waffling}. However, we found that these independently modelled prompts exhibit quality variations, leading to a larger semantic gap between text and image modalities and suboptimal classifiers.
To quantify this issue, we use the similarity between prompts and image prototypical features to evaluate the prompt quality, identifying the prompt with the lowest similarity as the "worst".
Taking CuPL \citep{prattWhatDoesPlatypus2023} as an example, the bottom density plot of Fig. \ref{Illustration} shows that the worst prompts selected from the entire set perform significantly worse than the others. Notably, the best-selected prompt is weaker than the zero-shot prompt obtained by averaging all prompts, indicating that classification performance improves with the inclusion of more prompts. This demonstrates that suboptimal prompts often arise because multiple independently obtained prompts in the training-free condition may not all suit downstream tasks. Without the guidance of image modality, suboptimal prompts limit the final classification performance. 
Here, we raise a question: \textit{Could the image be well utilized to mitigate the varying quality of text generated prompts?}

To jointly solve the 2 aforementioned issues, we explore the explicitly mutual guidance of images and text. 
To address the first issue (\emph{i.e.,} anomalous match), we observe that textual representation is generally more robust than image representation and that the knowledge from the two modalities is usually complementary. We consider incorporating textual features into the construction of the image classifier to alleviate the anomalous match of image modality.
To address the second issue (\emph{i.e.,} varying quality of prompts), we find that the suitability of prompts for specific tasks can be effectively assessed using image representations. Thus, we can use the image modality to guide a linear combination among prompts, enhancing the knowledge of text modality. Ultimately, we propose \textbf{T}ext-\textbf{I}mage \textbf{M}utual guidance \textbf{O}ptimization (TIMO) to address these two issues together.

\begin{figure}[t]
 \centering
    \includegraphics[width=1\columnwidth]{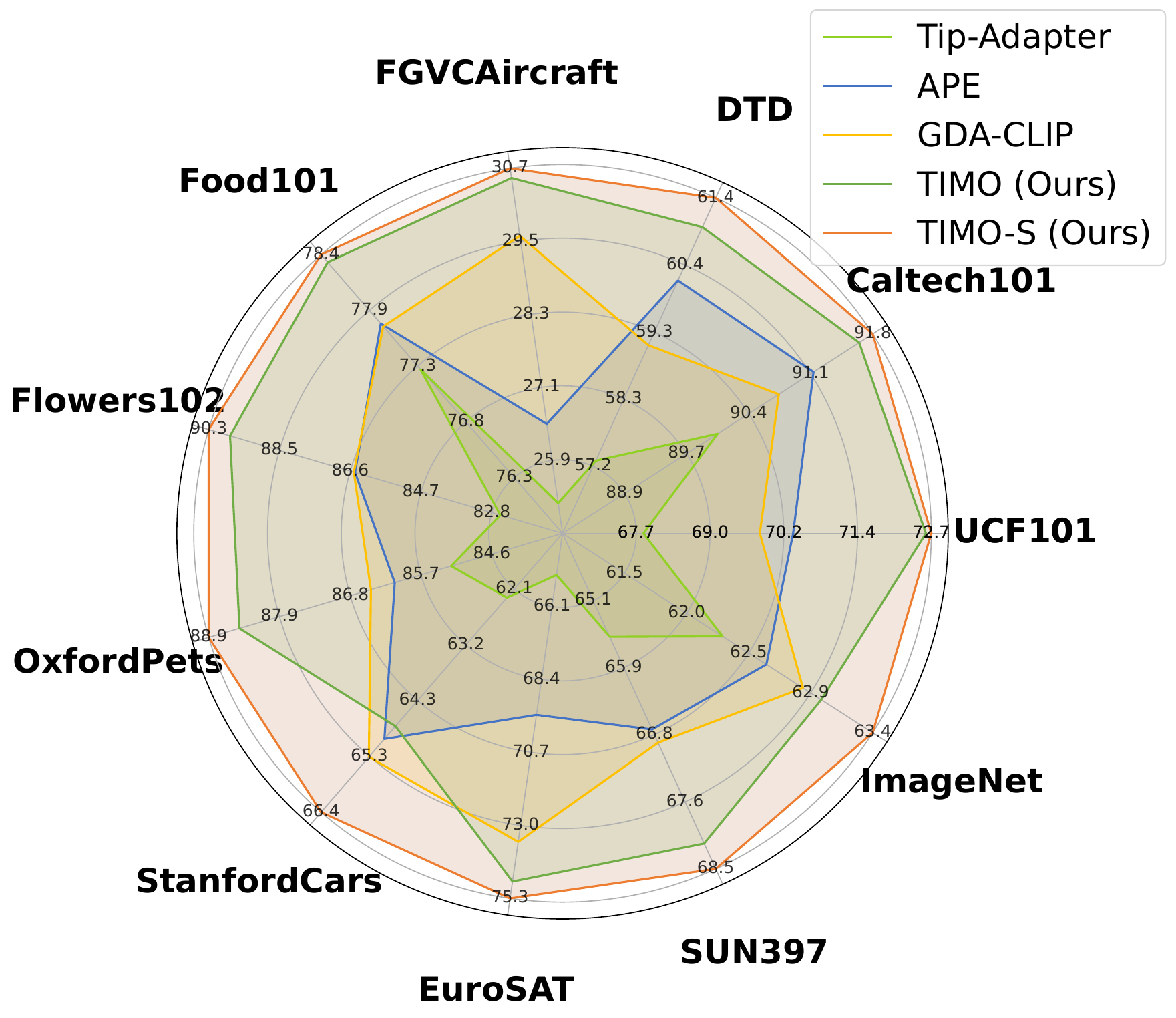}
    \caption{Comparison of the average Top-1 accuracy of various FSL methods across 11 datasets. TIMO-S outperforms the previous best method by an average of 1.76\%.} 
    \label{RadioPerformance}
\end{figure}

Our method can integrate with previous training-free methods, enhancing modality interaction. Combining Text-Guided-Image (TGI) and Image-Guided-Text (IGT) with Tip-Adapter (Tip-MG) improves its performance by 1.20\% across 11 datasets. Combining them with GDA-CLIP \citep{Wang_2024_Hard-to-Beat} (TIMO) achieves a 1.32\% improvement. Our enhanced variant, TIMO-S,
outperforms the second-best training-free method by 1.76\%. It even surpasses the best training-required method by 0.33\% while reducing time cost by approximately ×100. Performance comparisons of different training-free methods in Fig. \ref{RadioPerformance} show our method achieves the best results across all datasets. The research contributions of this paper are summarized as follows:

1) We recognize that independently modelling different modalities leads to severe anomalous match and varying prompt quality.

2) From the perspective of text-guided-image, we employ an instance-based transfer method to mitigate the severe anomalous matches in the image modality.

3) From the perspective of image-guided-text, we model an optimization task for the linear combination of prompts to address varying prompt quality in the text modality.

\section{Related Work}
\textbf{CLIP-based Multimodal Fusion}.
Vision-Language Models (VLMs), such as CLIP, have emerged as a new paradigm for foundational models by seamlessly integrating visual and textual information processing. In tasks like image classification, VLMs transfer textual knowledge to the visual modality to enhance performance. Previous training-free FSL methods like CLIP-Adapter \citep{Gao_2021_CLIP-Adapter}, Tip-Adapter \citep{Zhang_2022_Tip-Adapter}, APE \citep{Zhu_2023_Not}, and GDA-CLIP \citep{Wang_2024_Hard-to-Beat} have generally overlooked the importance of inter-modal interaction, resulting in limited knowledge transfer between modalities and consequently restricting the performance on downstream tasks.

Some works have recognized the importance of inter-modal interactions. Tip-X \citep{Udandarao_2023_SuS-X} uses inter-modal distances to calibrate intra-modal distances. VT-CLIP \citep{Qiu_2023_VT-CLIP} establishes semantic correlations between text and images through image-guided attention. CALIP \citep{guoCALIPZeroShotEnhancement2023} employs a parameter-free attention module to improve zero-shot classification. Cross-Modal \citep{Lin_2023_Multimodality} repurposes class names as additional support samples during the training stage to combine knowledge at the classification head. However, these methods either ignore that multiple prompts can represent the same class or lack in-depth interaction with multimodal fusion. We solve this by explicitly image-text mutual guidance and ultimately improving performance.

\textbf{Prompts Synthesis}.
VLMs have excellent performance in zero-shot classification, but recent studies show that task-specific prompt adjustments could improve performance \citep{Liu_2023_Pre-train, Huang_2022_Unsupervised}. Methods like CoOp \citep{Zhou_2022_Learning}, CoCoOp \citep{Zhou_2022_Conditionala}, ProDA \citep{Lu_2022_Prompt}, PLOT \citep{Chen_2023_PLOT}, and CPL \citep{Zhang_2024_Concept-Guideda} learn better prompts via gradient descent. Alternatively, studies reveal that prompts from LLMs contain crucial discriminatory characteristics of image classification tasks \citep{prattWhatDoesPlatypus2023}. CuPL \citep{prattWhatDoesPlatypus2023} and DCLIP \citep{Menon_2022_Visual} utilise LLMs to acquire fine-grained prompts, while WaffleCLIP \citep{Roth_2023_Waffling} suggests LLM-generated semantic descriptors enhance classification in a complementary way. Additionally, some works use other external resources like Wikipedia \citep{tahmasebzadehFewShotEventClassification2024, Paz-Argaman_2020_ZEST, Naeem_2022_I2DFormer} and WordNet \citep{Fellbaum_1998_WordNet, Roth_2022_Integrating, Shen_2022_K-LITE} to help prompts construction.

This paper finds that prompts generated by LLMs may not be fully exploited in FSL due to the modality gap with the image. Therefore, we derive prompt weights by modelling the optimization task of linear combination among prompts, obtaining rectified textual feature representations. This enhancement bridges the gap between different modalities, thereby improving image classification.

\begin{figure*}[t]
  \centering
  \includegraphics[width=1\linewidth]{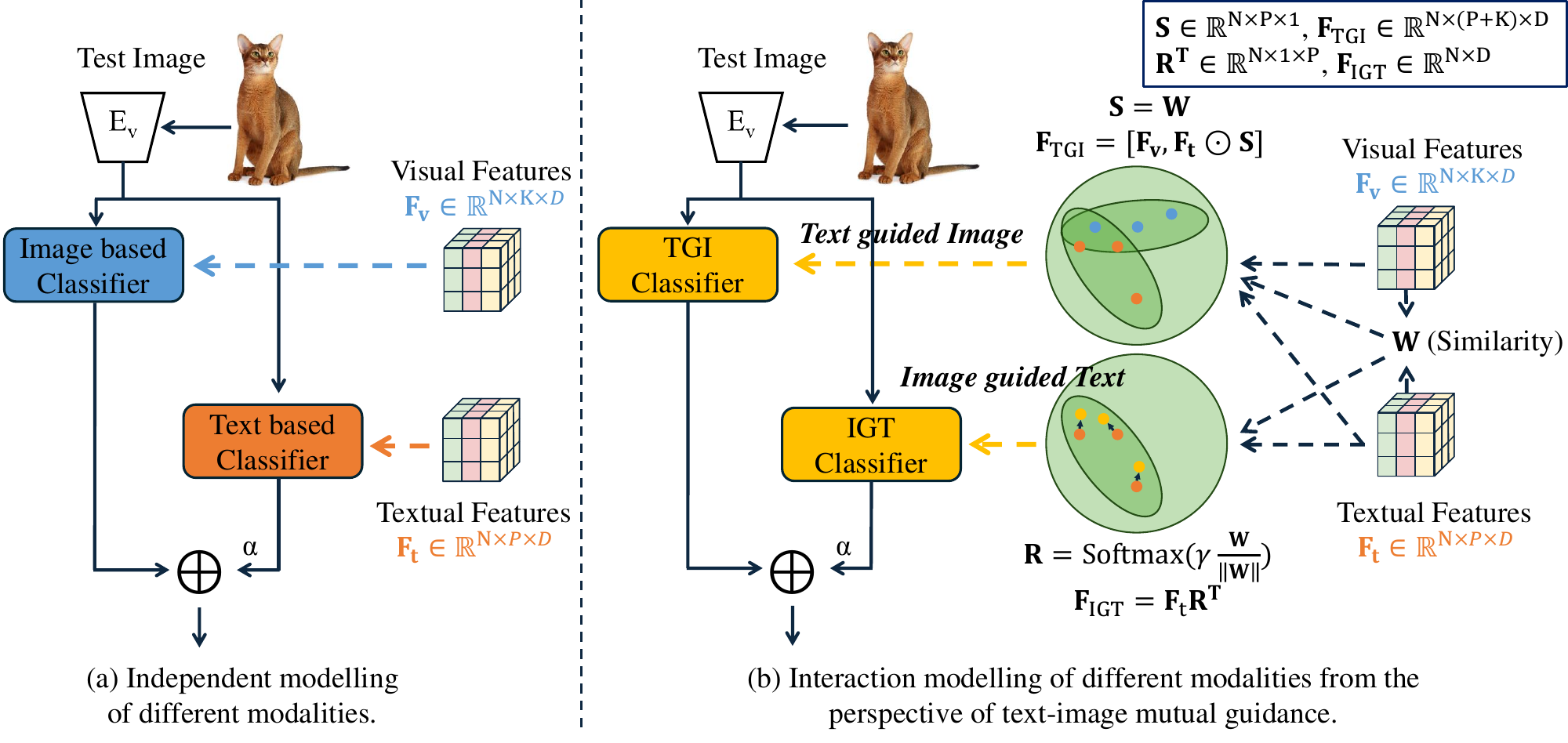}
  \caption{Comparison between existing CLIP-based few-shot methods and ours. (a) indicates the general architecture of previous work. (b) shows how text-image mutual guidance is applied in this work to interact deeply with different modalities.} 
  \label{Overview}
\end{figure*}

\section{Method} \label{Method}
We introduce the proposed technical details of image-text mutual guidance in CLIP-based few-shot classification. Firstly, we formulate the problem and establish relevant notations. Subsequently, two subsections elucidate how Image-Guide-Text (IGT) and Text-Guide-Image (TGI) are implemented, respectively. Then, we provide a detailed explanation of how TIMO and TIMO-S are constructed.

\subsection{Preliminaries} \label{Preliminaries}
We begin with a brief introduction to CLIP \citep{Radford_2021_Learning}, which is trained on paired text and image data, enabling it to map features from different modalities into a joint feature space. 
In the zero-shot setting, CLIP encodes the text with prompt templates (\emph{e.g.,} “a photo of a [CLASS]”) into the text feature 
$\mathbf{F_t} = \left[\mathbf{F}_\mathrm{t}^1,\ldots,\mathbf{F}_\mathrm{t}^i,\ldots,\mathbf{F}_\mathrm{t}^N \right] \in \mathbb{R}^{N \times P\times D} $, where $i$ represents $i$-th of the $N$ categories, $P$ is the number of prompts per category and $D$ is the feature dimension. 
Then following \citet{Snell_2017_Prototypical}, we average all $P$ prompts yields prototypical features $\mathbf{W_t} \in \mathbb{R}^{N \times D}$. 
Similarly, CLIP encodes the support images into the feature $\mathbf{F_v} \in \mathbb{R}^{N\times K \times D}$ with prototypical features $ \mathbf{W_v} = \left[\mathbf{W}_\mathrm{v}^1,\ldots,\mathbf{W}_\mathrm{v}^i,\ldots,\mathbf{W}_\mathrm{v}^N \right] \in \mathbb{R}^{N \times D} $, where $K$ is the number of shots in FSL. 
For a query image, it is encoded by CLIP as a visual feature $\mathbf{f_v} \in \mathbb{R}^{D}$.

In zero-shot classification, CLIP uses cosine similarity to build the zero-shot classifier and calculates the $\mathrm{logits}$, which denotes the prediction probabilities for all $N$ categories.

\begin{equation}
  \mathrm{logits}_\mathrm{t} = \frac{\mathbf{W}_\mathrm{t} \mathbf{f}_\mathrm{v}}{\left \Vert \mathbf{W}_\mathrm{t} \right \Vert \left \Vert \mathbf{f}_\mathrm{v} \right \Vert} \in \mathbb{R}^{N}.
  \label{zero-shot classification}
\end{equation}

In FSL, support image samples are often used in various ways to construct an additional classifier (\emph{i.e.,} $\operatorname{Classifer_{Image}})$ \citep{Zhang_2022_Tip-Adapter, Wang_2024_Hard-to-Beat}, independent of the aforementioned zero-shot classifier. Then, predictions from both image and text classifiers are combined by weighted summation for the final prediction. This process can be represented as:

\begin{equation}
        \mathrm{logits}_\mathrm{v} = \operatorname{Classifer_{Image}}(\mathbf{f}_\mathrm{v}) \in \mathbb{R}^{N},
\end{equation}
\begin{equation}
        \mathrm{logits}_\mathrm{o} = \mathrm{logits}_\mathrm{v} + \alpha \mathrm{logits}_\mathrm{t},
\end{equation}
where the hyperparameter $\alpha$ balances the weights of image and text predictions. However, previous SOTA methods build classifiers independently for each modality, ignoring inter-modal correlations. This oversight leads to a semantic gap and leaves single-modality weaknesses unaddressed. We propose a mutual guidance method between image and text modalities, detailed in the following subsection.

\subsection{Text-Guide-Image (TGI)} \label{TGI}
To fully exploit the interrelation between multimodality and alleviate the weakness of each single modality. Here, we discuss the details of the TGI classifier. 
Since CLIP encodes image and text modalities into homogeneous features located in the same feature space but with different data distributions, we propose using an instance-based transfer learning method. 
This involves computing a weight matrix $\mathbf{s}^i \in \mathbb{R}^{P}$ for each prompt with label $i$, and subsequently reordering $\mathbf{F}_t^i$ based on the descending sequence of $\mathbf{s}^i$. Here, we introduce a hyper-parameter $\beta$ to adjust the degree of text guidance for the image modality by controlling the number of text prompts involved in the instance-based transfer. The default value for $\beta$ is set to $P$, but by applying a grid search on the validation set, we can optimize this parameter, leading to our TIMO-S variant which will be discussed later.

\begin{equation}
    \mathbf{s}^i = \frac{\mathbf{F}_\mathrm{t}^i \mathbf{W}_\mathrm{v}^i}{\left \Vert \mathbf{F}_\mathrm{t}^i \right \Vert \left \Vert \mathbf{W}_\mathrm{v}^i \right \Vert} \in \mathbb{R}^{P},
\end{equation}
\begin{equation}
    \mathbf{s}^i = \operatorname{diag}(\mathbf{I}_\beta, \mathbf{0}_{P-\beta, P-\beta}) \mathbf{s}^i \in \mathbb{R}^{P}.
\end{equation}

Here, $\operatorname{diag}$ is defined as the operation that combines multiple block matrices into a diagonal matrix. Then the obtained weights $\mathbf{S} = \left[\mathbf{s}^1,\ldots,\mathbf{s}^i,\ldots,\mathbf{s}^N \right] \in \mathbb{R}^{N \times P}$ are element-wise multiplied with the textual features. Subsequently, the $\operatorname{Concat}$ operation combines the image features with the non-zero weighted text features to achieve instance-based transfer. 
This results in the feature $\mathbf{F}_\mathrm{TGI}$, which can be used to construct the TGI classifier.

\begin{equation}
    \mathbf{F}_\mathrm{TGI} = \operatorname{Concat}(\mathbf{F}_\mathrm{v}, \mathbf{F}_\mathrm{t} \odot \mathbf{S}) \in \mathbb{R}^{N \times (K+\beta) \times D}.
    \label{F_TGI}
\end{equation}

After obtaining the refined features $\mathbf{F}_\mathrm{TGI}$, we use the previously established image classifier construction method (\emph{i.e.,} $\operatorname{ImageClassifierBuilder}$) to build the classifier $\operatorname{Classifier_{TGI}}$. By substituting $\operatorname{ImageClassifierBuilder}$, our method can achieve the plug-and-play capability with previous SOTA methods. Then, using the $\operatorname{Classifier_{TGI}}$, we obtain the logits corresponding to the query image feature $\mathbf{f}_\mathrm{v}$.

\begin{equation}
    \mathrm{logits}_\mathrm{{TGI}} = \operatorname{Classifier_{TGI}}(\mathbf{f}_\mathrm{v}) \in \mathbb{R}^{N}.
    \label{TGIClassifier}
\end{equation}

\subsection{Image-Guide-Text (IGT)} \label{IGT}
After the discussion of TGI, we now further discuss how to implement an image-guide-text (IGT) classifier to rectify biased prompts that are independently obtained. From the perspective of ensemble learning, a good IGT classifier needs to satisfy two targets \citep{Krogh_1994_Neural}. 
\textbf{Target $\text{\uppercase\expandafter{\romannumeral1}}$}: the IGT Classifier should classify the data correctly; 
\textbf{Target $\text{\uppercase\expandafter{\romannumeral2}}$}: the IGT classifier should hold the individuality information (\emph{i.e.,}, keep diversity) from the TGI classifier.

For computational convenience, we first normalize all features using the L2 norm. To achieve \textbf{Target $\text{\uppercase\expandafter{\romannumeral1}}$} in a training-free setting, we define a vector $\mathbf{r}^i \in \mathbb{R}^{P}$ to maximize the multiplication result between the transformed matrix $\mathbf{r}^{i\top} \mathbf{F}_t^i$ and the corresponding prototype visual features $\mathbf{W}_v^i$ in category $i$. This operation computes the cosine similarity between each vector in $\mathbf{F}_t^i$ and the prototype $\mathbf{W}_v^i$, with $\mathbf{r}^i$ assigning higher weights to more similar vectors. However, the vector's magnitude also affects the optimisation objective. Thus, to meet \textbf{Target $\text{\uppercase\expandafter{\romannumeral2}}$}'s requirements, we constrain the norm of $\mathbf{r}^i$ to prevent excessive loss of original text features in the transformed matrix $\mathbf{r}^{i\top} \mathbf{F}_t^i$. Here, we introduce a hyperparameter $\gamma$ to adjust the degree of image guidance for the text modality by controlling the norm of $\mathbf{r}^i$. The final optimization task is formulated as follows:

\begin{equation}
    \begin{aligned}
    \max\limits_{\mathbf{r}^{i}} \mathbf{r}^{i\top} \mathbf{F}_\mathrm{t}^i \mathbf{W}_\mathrm{v}^i, \qquad
    \mathrm{s.t.} \left \Vert \mathbf{r}^i \right \Vert = \gamma.
    \end{aligned} \label{new_opt_task}
\end{equation}

Using the Lagrange multipliers method, Eq. (\ref{new_opt_task}) can be transformed into an unconstrained problem:
\begin{equation}
    \min\limits_{\mathbf{r}^i} \mathcal{L} =  -\mathbf{r}^{i\top} \mathbf{F}_\mathrm{t}^i \mathbf{W}_\mathrm{v}^i + \lambda (\mathbf{r}^{i\top}	\mathbf{r}^i - \gamma).
\end{equation}

By solving the system of two equations where the partial derivatives $\triangledown_{\mathbf{r}^i} \mathcal{L}$ and $\triangledown_\lambda \mathcal{L}$ are both equal to zero, we obtain two sets of solutions.
\begin{equation}
    \left\{ 
    \begin{array}{cc}
        \begin{aligned}
            \lambda &= \pm \frac{1}{2\gamma} \left \Vert \mathbf{F}_\mathrm{t}^i \mathbf{W}_\mathrm{v}^i \right \Vert, \\
            \mathbf{r}^i &= \pm \frac{ \gamma \mathbf{F}_\mathrm{t}^i \mathbf{W}_\mathrm{v}^i}{\left\Vert \mathbf{F}_\mathrm{t}^i \mathbf{W}_\mathrm{v}^i \right \Vert}.
        \end{aligned} 
    \end{array}
    \right.
\end{equation}

Substituting $\lambda$ back into the original equation, when $\lambda$ is positive, the original function with respect to $\mathbf{r}^i$ is convex, yielding the minimum value of the function $\mathcal{L}$. Otherwise, the maximum $\mathcal{L}$ is obtained. Hence, the combination $\lambda = \frac{1}{2\gamma} \left \Vert \mathbf{F}_\mathrm{t}^i \mathbf{W}_\mathrm{v}^i  \right \Vert$ and $\mathbf{r}^i = \frac{ \gamma \mathbf{F}_\mathrm{t}^i \mathbf{W}_\mathrm{v}^i}{\left\Vert \mathbf{F}_\mathrm{t}^i \mathbf{W}_\mathrm{v}^i \right \Vert}$ is chosen. Since both $\mathbf{F}_\mathrm{t}^i$ and $\mathbf{W}_\mathrm{v}^i$ are pre-normalised, $\mathbf{r}^i$ is essentially equivalent to the normalisation of $\mathbf{s}^i$ (discussed in the previous section). In the end, the $\mathrm{logits}_\mathrm{IGT}$ are defined as below:

\begin{equation}
    \mathbf{R} = \left[\ldots,\operatorname{SoftMax}( \mathbf{r}^i),\ldots \right] \in \mathbb{R}^{N \times P},
\end{equation}
\begin{equation}
    \mathbf{F}_\mathrm{IGT} = \mathbf{F}_\mathrm{t} \mathbf{R^\top} \in \mathbb{R}^{N \times D}. 
    \label{F_IGT}
\end{equation}

Similar to the construction of $\operatorname{Classifier_{TGI}}$, we can use the $\mathbf{F}_\mathrm{IGT}$ and $\operatorname{TextClassifierBuilder}$ to construct a classifier $\operatorname{Classifier_{IGT}}$ and then achieve the prediction of $\mathbf{f}_\mathrm{v}$.
\begin{equation}
    \mathrm{logits}_\mathrm{IGT} = \operatorname{Classifier_{IGT}}(\mathbf{f}_\mathrm{v}). 
    \label{IGTClassifier}
\end{equation}

\subsection{Image-Text Mutual Guidance} \label{Image-Text Mutual Guidance}
As aforementioned, TGI and IGT guide the image and text modalities separately. Additionally, their ability to seamlessly integrate with existing methods (\emph{i.e.,} plug-and-play) enables them to effectively enhance the performance of previous approaches. We use the latest and top-performing method, GDA-CLIP \citep{Wang_2024_Hard-to-Beat}, as an example to illustrate how TGI and IGT can be integrated (Note that TGI and IGT can easily combine with other methods as well), leading to the proposed TIMO and TIMO-S.

In general, as shown in Fig. \ref{Overview}, previous methods lack an Image-Text Mutual Guidance module and thus construct independent classifiers using the encoded image features $\mathbf{F}_\mathrm{v}$ and text features $\mathbf{F}_\mathrm{t}$. We found this independent modelling leads to suboptimal classifiers. To address this, we introduced TGI and IGT components, transforming the original features into refined features $\mathbf{F}_\mathrm{TGI}$ and $\mathbf{F}_\mathrm{IGT}$ via Eq. (\ref{F_TGI}) and (\ref{F_IGT}). Replacing the original unimodal features in GDA-CLIP with our refined features, while keeping the classifier construction process unchanged, allows for easy integration of our method with GDA-CLIP or other methods. 

By incorporating TGI and IGT into the previous methods, we proposed TIMO. And the final prediction of TIMO can be obtained by ensemble $\mathrm{logits}_\mathrm{TGI}$ from Eq. (\ref{TGIClassifier}) and $\mathrm{logits}_\mathrm{IGT}$ from Eq. (\ref{IGTClassifier}).
\begin{equation}
    \mathrm{logits} = \mathrm{logits}_\mathrm{TGI} + \alpha \mathrm{logits}_\mathrm{IGT},
\end{equation}
where $\alpha\in \{10^{-4},10^{-3},\ldots,10^{4}\}$ and we use grid search, similar to previous methods, on the validation set to determine the optimal value of $\alpha$. Ultimately, by incorporating TGI and IGT, we achieved mutual guidance between image and text modalities, which not only mitigated the inherent weaknesses of individual modalities but also reduced the semantic gap between them.

\textbf{Construction of TIMO-S}. Ablation experiments reveal that adjusting the hyperparameters $\beta$ and $\gamma$, which control the degree of mutual guidance, can further enhance TIMO's performance. Therefore, we propose to conduct an additional grid search on the validation set for $\beta$ and $\gamma$, resulting in a more robust model, TIMO-S. Note that, all compared training-free methods employ similar grid searches, ensuring fair comparison.

\section{Experiments} \label{Experiments}
\subsection{Setting and Implementation} \label{Setting and Implementation}
\textbf{Dataset}.
According to previous works \citep{Wang_2024_Hard-to-Beat, Zhang_2022_Tip-Adapter}, we use 11 image classification benchmarks to evaluate few-shot image classification. The specific datasets and their licenses are detailed in the Appendix.

\textbf{Baselines}.
Our method is compared with 2 categories of existing methods. 1) For training-free methods, Zero-Shot CLIP \citep{Radford_2021_Learning}, Tip-Adapter \citep{Zhang_2022_Tip-Adapter}, APE \citep{Zhu_2023_Not} and GDA-CLIP \citep{Wang_2024_Hard-to-Beat} are compared; 2) For training-required methods, Tip-Adapter-F \citep{Zhang_2022_Tip-Adapter}, Cross-Modal \citep{Lin_2023_Multimodality} APE-T \citep{Zhu_2023_Not} are compared.

\textbf{Implementation Detail}.
This paper follows the data preprocessing protocol of CoOp \citep{Zhou_2022_Learning} and Tip-Adapter \citep{Zhang_2022_Tip-Adapter}. For TIMO, we set the hyper-parameter $\gamma$ to $50$ for all datasets, except $1$ for ImageNet and $100$ for Flowers102. To evaluate our method, we follow \citep{Zhu_2023_Not, Wang_2024_Hard-to-Beat, Zhang_2022_Tip-Adapter} using ResNet-50 (RN50) \citep{He_2016_Deep} as the default visual encoder of CLIP, compare 1, 2, 4, 8, and 16 few-shot training sets, and use textual prompts from APE \citep{Zhu_2023_Not} and CuPL \citep{prattWhatDoesPlatypus2023}. All experiments use PyTorch \citep{Paszke_2019_PyTorch} and are conducted on a single NVIDIA GeForce RTX 3090Ti GPU.

\begin{table*}[h]
\centering
\footnotesize
\setlength{\tabcolsep}{3.5mm}{
    \begin{tabular}{*{8}{c}}
       \toprule
       \multirow{2}*{Methods} & \multirow{2}*{Training-Free} & \multicolumn{5}{c}{Number of shots} & \multirow{2}*{Avg.} \\
       \cmidrule(lr){3-7} 
       & & 1 & 2 & 4 & 8 & 16 \\
       \midrule
       Linear Probe CLIP$_\text{ ICML'21}$ \citep{Radford_2021_Learning} &\XSolidBrush    & 36.10 & 46.99 & 56.72 & 64.66 & 70.56 & 55.01 \\
       CoOp$_\text{ IJCV'22}$ \citep{Zhou_2022_Learning} &\XSolidBrush    & 59.56 & 61.78 & 66.47 & 69.85 & 73.33 & 66.20 \\
       Tip-Adapter-F$_\text{ ECCV'22}$ \citep{Zhang_2022_Tip-Adapter}       & \XSolidBrush  & 65.96 & 67.86 & 70.28 & 72.99 & 75.87 & 70.59 \\
       Cross-Modal Partial$_\text{ CVPR'23}$ \citep{Lin_2023_Multimodality} & \XSolidBrush  & 65.97 & 68.37 & 71.14 & \underline{73.96} & \textbf{77.05} & \underline{71.30} \\
       APE-T$_\text{ ICCV'23}$ \citep{Zhu_2023_Not}               & \XSolidBrush  & 65.96 & 68.56 & 71.27 & 73.88 & 76.69 & 71.27 \\
       LP++$_\text{ CVPR'24}$ \citep{Huang_2024_LP} &\XSolidBrush    & 64.56 & 66.32 & 69.41 & 71.79 & 74.70 & 69.36 \\
       CLAP$_\text{ CVPR'24}$ \citep{Silva-Rodriguez_2024_Closer} &\XSolidBrush    & 66.10 & \textbf{69.10} & \textbf{72.10} & 73.50 & 75.10 & 71.18 \\
       \midrule
       Zero-Shot CLIP$_\text{ ICML'21}$ \citep{Radford_2021_Learning} &\CheckmarkBold    & 62.11 & 62.11 & 62.11 & 62.11 & 62.11 & 62.11 \\
       Tip-Adapter$_\text{ ECCV'22}$ \citep{Zhang_2022_Tip-Adapter}         &\CheckmarkBold & 64.24 & 65.51 & 67.31 & 69.00 & 70.61 & 67.33 \\
       Tip-MG (Ours)         &\CheckmarkBold & 66.14 & 67.35 & 68.86 & 70.58 & 71.66 & 68.92 \\
       APE$_\text{ ICCV'23}$\citep{Zhu_2023_Not}                 &\CheckmarkBold & 65.78 & 67.24 & 69.32 & 71.16 & 72.99 & 69.30 \\
       GDA-CLIP$_\text{ ICLR'24}$ \citep{Wang_2024_Hard-to-Beat}            &\CheckmarkBold & 63.69 & 66.39 & 69.86 & 73.17 & 76.23 & 69.87 \\
       \rowcolor{gray!40}  
       TIMO (Ours)         &\CheckmarkBold & \underline{66.43} & 68.15 & 71.20 & 73.55 & 76.61 & 71.19 \\
       \rowcolor{gray!40}  
       TIMO-S (Ours)       &\CheckmarkBold & \textbf{66.78} & \underline{68.69} & \underline{71.63} & \textbf{74.20} & \underline{76.86} & \textbf{71.63} \\
       \bottomrule
    \end{tabular}
    \caption{Performance (\%) of few-shot classification on 11 datasets, showing top-1 accuracy across 3 random seeds. TIMO surpasses all previous training-free approaches. Furthermore, its enhanced variant, TIMO-S, outperforms all previous training-required methods. Results of Linear Probe CLIP and CLAP are sourced from \citet{Huang_2024_LP} and \citet{Wang_2024_Hard-to-Beat}.} \label{PerformanceTable}
}
\end{table*}

\subsection{Main Result}
\textbf{Performance Analysis}.
Tab. \ref{PerformanceTable} shows few-shot classification results for both training-free and training-required methods, with bolded and underlined results representing optimal and suboptimal results separately. 
Linear Probe CLIP \citep{Radford_2021_Learning} simply fine-tunes the classifier but results in performance degradation in low-shot settings than zero-shot CLIP \citep{Radford_2021_Learning}. 
Subsequent works strive to preserve the model's zero-shot capabilities, where our TIMO enhance and spreads these capabilities through cross-modal mutual guidance mechanism, surpassing all previous training-free methods (especially in low-shot settings).
Additionally, our enhanced variant, TIMO-S, even outperforms all training-required methods under training-free conditions.

\begin{figure}[h]
    \centering
    \includegraphics[width=1\columnwidth]{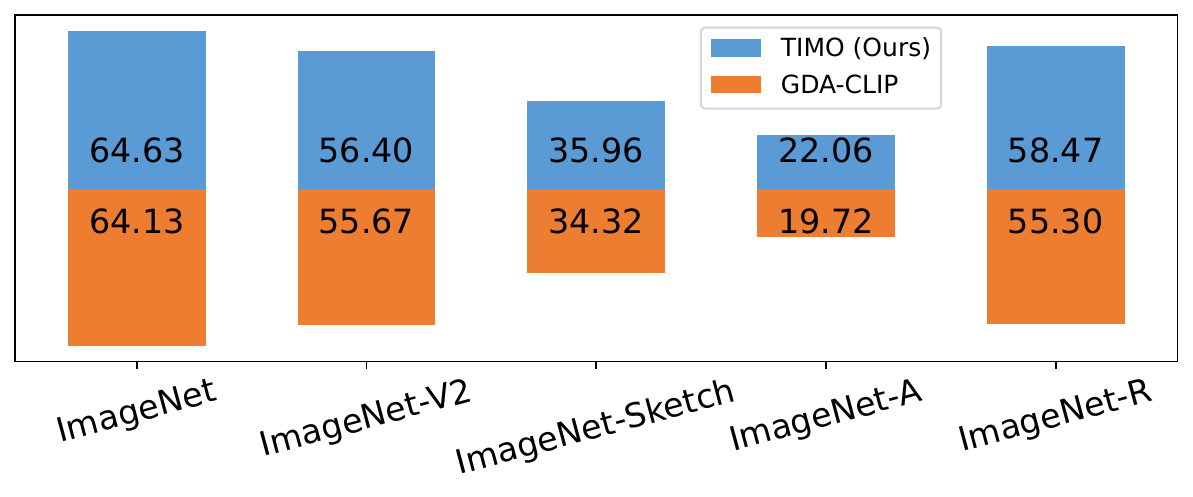}
    \caption{Domain Generalization Performance (\%) of TIMO and GDA-CLIP.} 
    \label{BarChartRobustness}
\end{figure}
\textbf{Robustness Analysis}. \label{Robustness Experiments}
To evaluate the robustness of our method under different conditions, we test its performance using different CLIP backbones. The results are presented in Tab. \ref{ArchitectureTable}, indicating that TIMO-S consistently outperforms prior methods across all visual encoders. Additionally, in line with previous research \citep{Wang_2024_Hard-to-Beat}, we use 16-shot ImageNet as the training data and test on out-of-distribution datasets ImageNet-V2, ImageNet-Sketch, ImageNet-A and ImageNet-R to evaluate the domain generalisation capability of our models based on RN50-based CLIP. Fig. \ref{BarChartRobustness} shows that TIMO-S outperforms GDA-CLIP across all target datasets. We believe this is due to the robust textual features being more fully engaged in the image classification task.

\begin{table*}
    \footnotesize
    \centering
    \setlength{\tabcolsep}{7.3mm} {
        \begin{tabular}{*{6}{c}}
           \toprule
           Methods & RN50 & RN101 & ViT-B/32 & ViT-B/16 & Avg. \\
           \midrule
           Tip-Adapter \citep{Zhang_2022_Tip-Adapter} & 67.33 & 68.13 & 70.07 & 74.20 & 69.92 \\
           APE \citep{Zhu_2023_Not}         & 69.30 & 70.41 & 71.78 & 75.89 & 71.84 \\
           GDA-CLIP \citep{Wang_2024_Hard-to-Beat}    & 69.87 & 71.60 & 72.70 & 76.69 & 72.71 \\
           \rowcolor{gray!40}  
           TIMO (Ours)      & \underline{71.19} & \underline{72.24} & \underline{73.71} & \underline{77.61} & \underline{73.69} \\
           \rowcolor{gray!40}  
           TIMO-S (Ours)    & \textbf{71.63} & \textbf{72.98} & \textbf{74.05} & \textbf{77.97} & \textbf{74.16} \\
           \bottomrule
        \end{tabular}
    }
    \caption{The average performance (\%) of few-shot classifications using different visual encoders across 11 datasets and 3 random seeds. } \label{ArchitectureTable}
\end{table*}

\subsection{Ablation Study} \label{Ablation Study}

\begin{figure}
    \centering
    \includegraphics[width=1\linewidth]{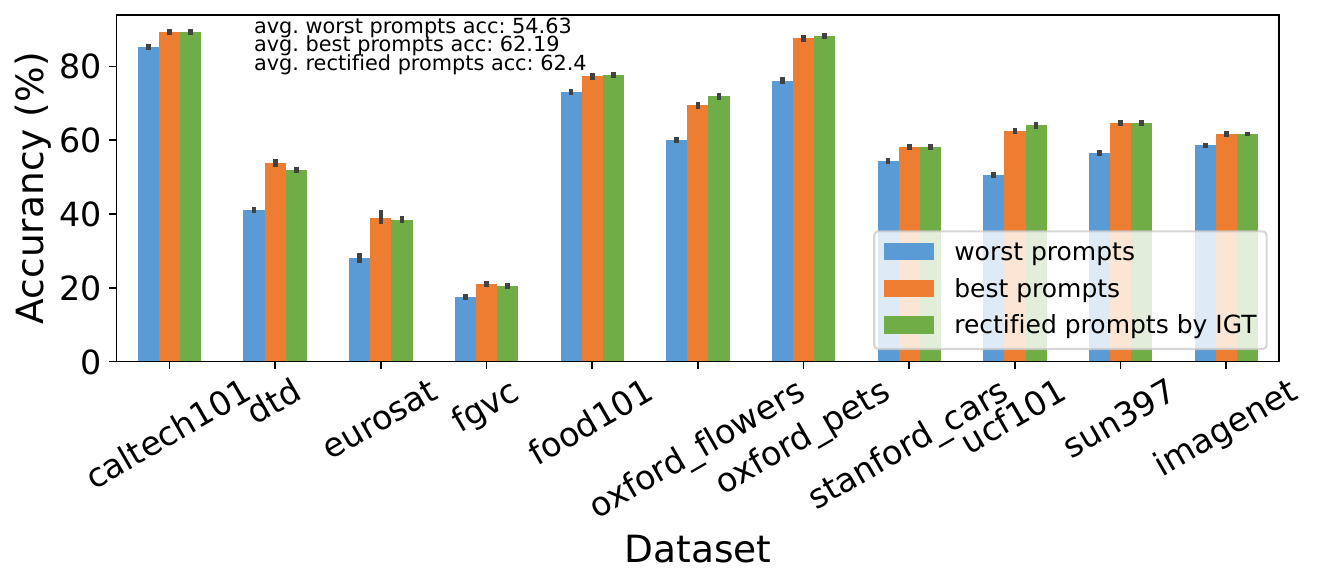}
    \caption{Ablation of prompt selection.} \label{BarChartPromptEffectiveness}
    \label{fig:enter-label}
\end{figure}

\begin{table}[h]
    \centering
    \footnotesize
    \begin{tabular}{*{5}{c}}
       \toprule
       Methods & TGI & IGT & Avg (\%). \\
       \midrule
       \multirow{4}*{Tip-Adapter \citep{Zhang_2022_Tip-Adapter}} 
        & ----- & ----- & 67.28 \\
       & +TGI & & 68.35 \\
       & & +IGT & 68.43 \\
       & +TGI & +IGT & 68.92 \\ 
       \midrule
       \specialrule{0em}{1.5pt}{1.5pt}
       \midrule
       \multirow{4}*{GDA-CLIP \citep{Wang_2024_Hard-to-Beat}} 
       & ----- & ----- & 69.87 \\
       & +TGI & & 70.62 \\
       & & +IGT & 70.65 \\
       & +TGI & +IGT & 71.19 \\
       \bottomrule
    \end{tabular}
    \caption{Ablation study on different modules and their combinations.} \label{AblationStudy}
\end{table}
\textbf{Effectiveness of TGI and IGT}.
We investigate whether the developed TGI and IGT classifiers effectively address the previously noted issues: anomalous match and varying quality of generated prompts.

For the first issue, as shown in the bottom of Fig. \ref{Illustration}, our method effectively reduces the number of anomalous matches compared to all previous approaches. Furthermore, our method's effectiveness has been extensively validated across multiple datasets through experiments.

For the second issue, we evaluated the performance using all prompts, only the best half of the prompts, only the worst half of the prompts, and rectified prompts by IGT, as shown in Fig. \ref{BarChartPromptEffectiveness}. The results indicate that our method achieves overall better task-relevant textual representations.

Additionally, we conducted ablation experiments on the performance of TGI and IGT classifiers and their combinations. Given the plug-and-play nature of our approach, Tab. \ref{AblationStudy} shows that integrating either TGI or IGT into previous methods enhances the model's discriminative ability. The best performance improvement is achieved when both TGI and IGT are applied together.

\textbf{Efficacy of hyperparameters.} Our method involves 3 hyperparameters. We conducted experiments on the DTD dataset to evaluate their impact on the model's performance.

\begin{figure}[htb]
    \centering
    \includegraphics[width=1\columnwidth]{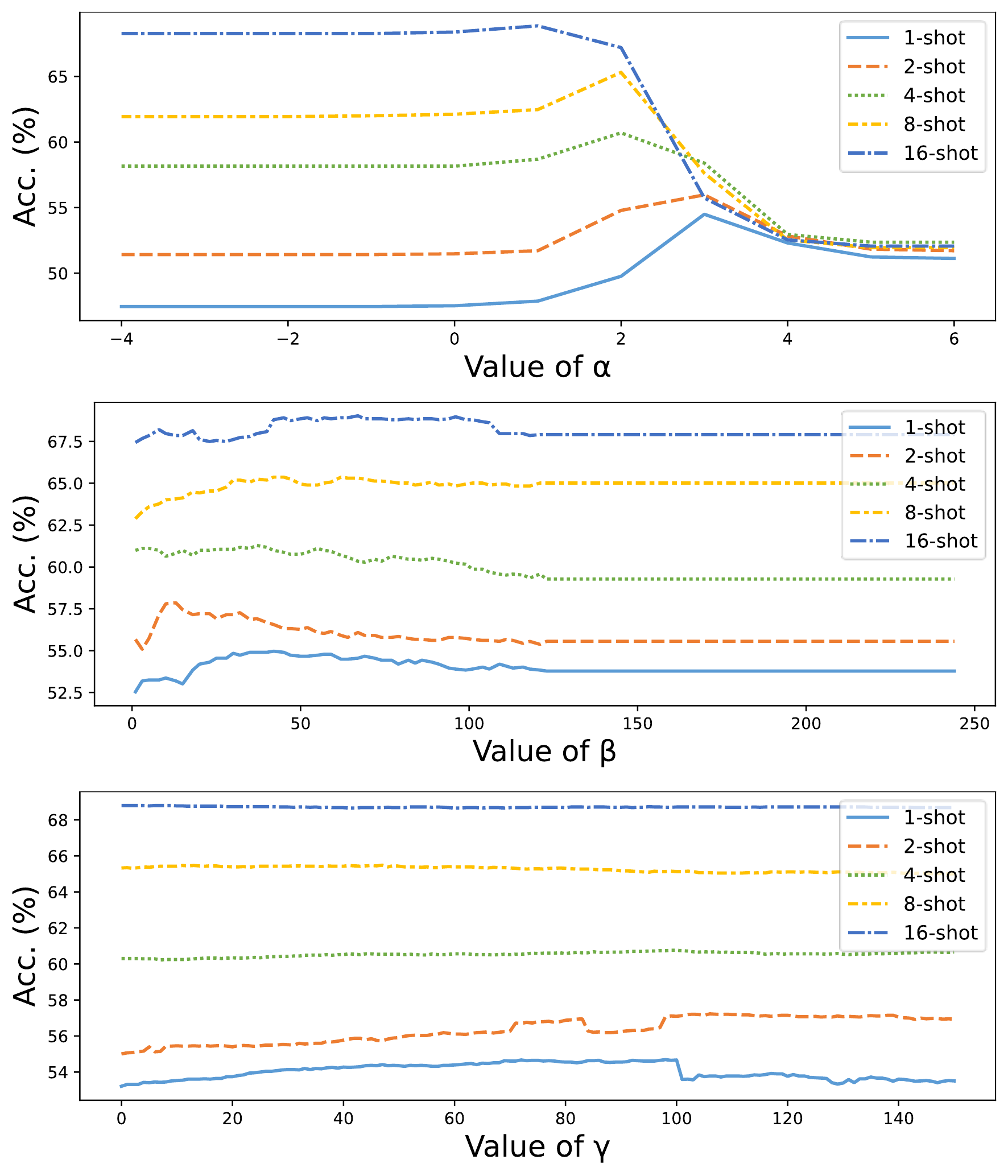}
    \caption{Ablation of hyperparameters.} 
    \label{HyperAbalation}
\end{figure}
$\alpha$ balances the weight of the TGI and IGT components. When $\alpha$ is 0, only TGI is active. As $\alpha$ approaches infinity, only IGT is active. Fig. \ref{HyperAbalation} shows the ablation experiments for $\alpha$, indicating that as $\alpha$ increases, performance under different shots starts to converge.

$\beta$ and $\gamma$ adjust the degree of text/image guidance for the image/text modality. As shown in Fig. \ref{HyperAbalation}, their overall effect on performance is relatively stable, but in rare cases, they may cause abrupt performance changes.

\textbf{Different Prompt Templates}.
We investigate how different prompt methods may affect the effectiveness of the IGT. In this regard, we primarily focus 4 prompt methods: CuPL \citep{prattWhatDoesPlatypus2023}, DCLIP \citep{Menon_2022_Visual} WaffleCLIP \citep{Roth_2023_Waffling} and use them all together.

\begin{figure}[!ht]
    \centering
    \includegraphics[width=1\columnwidth]{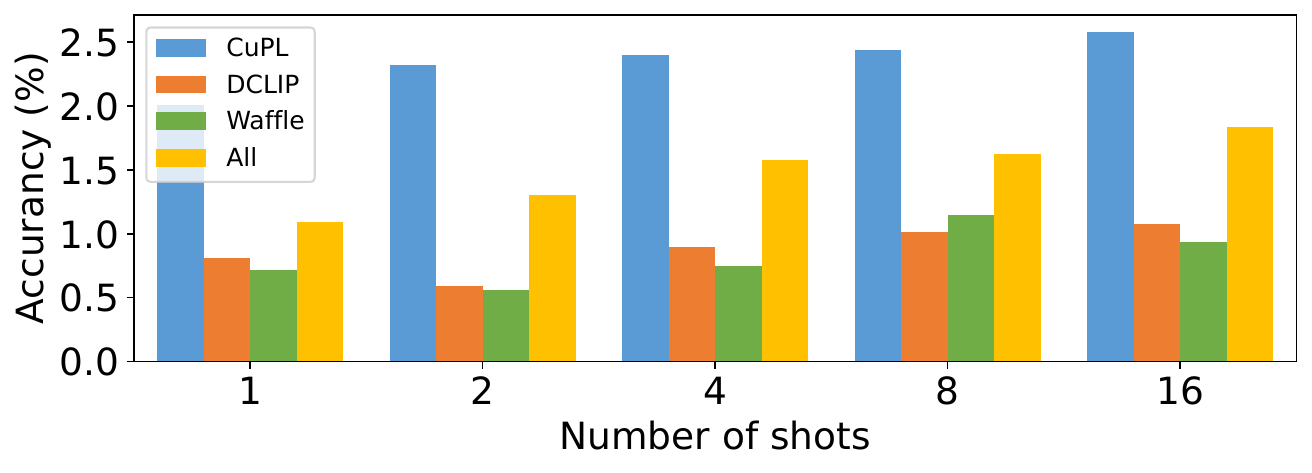}
    \caption{Performance improvement brought by IGT compared with zero-shot classification. } 
    \label{BarChartPromptAblation}
\end{figure}

From Fig. \ref{BarChartPromptAblation}, it is evident that different prompt templates impact the efficacy of the IGT classifier. Another interesting observation is that IGT improves more on CuPL than using all prompts. This indicates that the FSL could perform better when employing prompts containing richer and more diverse information, such as CuPL.

\subsection{Complexity Discussion} \label{Detailed Complexity Discussion}
\begin{table}[t]
    \centering
    \footnotesize
    \setlength{\tabcolsep}{1.5mm} {
        \begin{tabular}{*{4}{c}}
            \toprule
            Methods & Acc. (\%) & Time (s) & Param. \\
            \midrule
            Tip-Adapter \citep{Zhang_2022_Tip-Adapter} & 62.88 & 257.4 & $NKD$ \\
            APE \citep{Zhu_2023_Not}         & 63.43 & 787.5 & $NKD$ \\
            GDA-CLIP \citep{Wang_2024_Hard-to-Beat}    & 64.46 & 1.1 & $ND$ \\
            TIMO (Ours)        & 64.71 & 1.1  &$ND$\\
            \rowcolor{gray!40}  
            TIMO-S (Ours)      & 64.85 & 6.0  &$ND$\\
            \bottomrule
        \end{tabular}
    }
    \caption{Complexity Analysis on 16-shots ImageNet.}
    \label{ExpenseAnalysisTable}
\end{table}
From Tab. \ref{ExpenseAnalysisTable}, our method focuses solely on image-text mutual guidance, which does not introduce extra parameters or increase space complexity. Regarding time complexity, TIMO adds an O(1) operation compared to previous methods, maintaining the overall time complexity. However, TIMO includes additional hyperparameter search, resulting in better performance and relatively higher time cost.

\section{Conclusion} \label{Conclusion}
This paper introduces TIMO, a novel model for CLIP-based few-shot image classification. Unlike prior training-free methods that independently model different modalities, TIMO introduces a unique perspective by incorporating the IGT classifier and the TGI classifier to address the problem of the modality's inherent weakness and suboptimal prompt. Extensive experiments demonstrate the effectiveness and universality of the cross-modal mutual guidance, showing that TIMO-S outperforms previous best training-required methods under training-free conditions. 
However, our method does not employ novel training techniques to fully exploit CLIP's potential, and our multimodal mutual guidance mechanism could potentially be extended to other fields (\textit{e.g.}, image retrieval). These challenges remain open for future exploration.

\section{Acknowledgments}
This work was supported by the National Key R\&D Program of China (2023ZD0120700, 2023ZD0120701), NSFC Project (62222604, 62206052), China Postdoctoral Science Foundation (2024M750424), the Fundamental Research Funds for the Central Universities (020214380120), the State Key Laboratory Fund (ZZKT2024A14), the Postdoctoral Fellowship Program of CPSF (GZC20240252), and the Jiangsu Funding Program for Excellent Postdoctoral Talent (2024ZB242).

\bibliography{aaai25}

\clearpage

\input{appendix}

\end{document}

%% file: appendix.tex
\renewcommand{\thefigure}{S\arabic{figure}}
\renewcommand{\thetable}{S\arabic{table}}
\setcounter{figure}{0}
\setcounter{table}{0}
\appendix

\begin{figure*}[!h]
  \centering
  \includegraphics[width=1\linewidth]{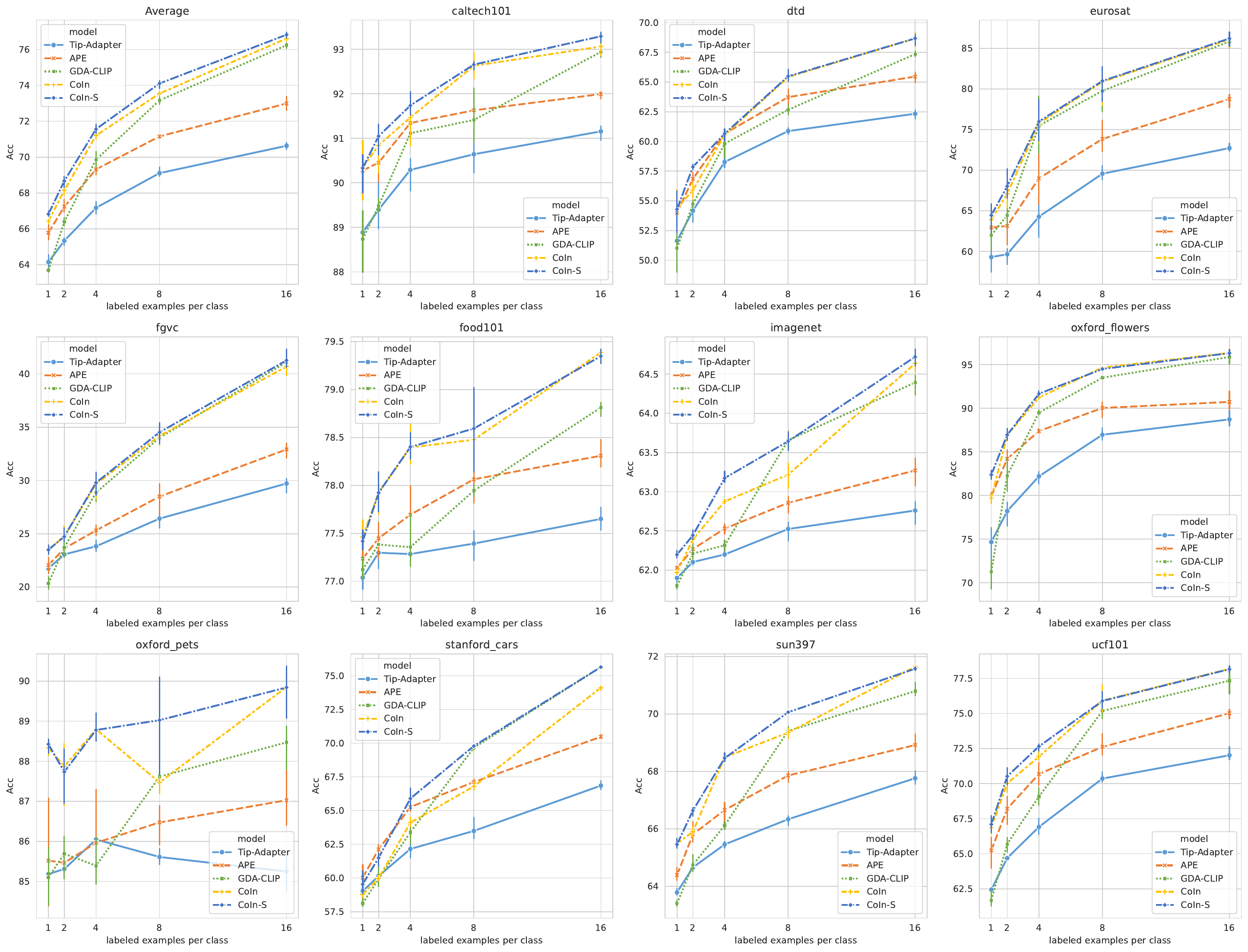}
  \caption{Few-shot classification accuracy (\%) of different models on 11 image classification datasets across 3 random seeds. Our best method, TIMO-S, surpasses previous approaches on nearly all datasets.} 
  \label{LineChartPerformance}
\end{figure*}

\subsection{A. Pseudo-code} 
The PyTorch style pseudocode of TIMO can be found at Algorithm \ref{alg:algorithm}.

\begin{algorithm}[htb]
    \caption{Main algorithm of TIMO.}
    \label{alg:algorithm}
    \textbf{Input}: Image features $\mathbf{F}_\mathrm{v} \in \mathbb{R}^{N \times K\times D}$, text features $\mathbf{F}_\mathrm{t} \in \mathbb{R}^{N \times P\times D}$, a test image features extracted by CLIP as $\mathbf{f}_\mathrm{v} \in \mathbb{R}^{D}$, ImageClassifierBuilder, TextClassifierBuilder.\\
    \textbf{Parameter}: The weight of multimodal branch $\alpha$, fusion degree controller $\gamma$.\\
    \textbf{Output}: Output prediction logits.
    \begin{algorithmic}[1] 
    
        \STATE \textcolor{gray}{\#Modeling}.
        \STATE $\mathbf{F}_\mathrm{vp} = \mathbf{F}_\mathrm{v}.\textrm{mean(1)}.\textrm{unsqueeze(1)}$.
        \STATE $\mathbf{F}_\mathrm{vp} =\mathbf{F}_\mathrm{vp} / \mathbf{F}_\mathrm{vp}.\textrm{norm}(-1, \textrm{keepdim=True})$.
        \STATE $\mathbf{W} = \textrm{CosineSimilarity}(\mathbf{F}_\mathrm{t}, \mathbf{F}_\mathrm{vp}, -1).$
        
        \STATE \textcolor{gray}{\#Building TGI Classifier}.
        \STATE $\mathbf{F}_\mathrm{TGI} = \textrm{concat}([\mathbf{F}_\mathrm{v}, \mathbf{F}_\mathrm{t}*\mathbf{W}.\textrm{unsqueeze}(-1)], 1).$
        \STATE $\mathrm{Classifier_{TGI}} = \textrm{ImageClassifierBuilder}(\mathbf{F}_\mathrm{TGI}).$
        
        \STATE \textcolor{gray}{\#Building IGT Classifier}.
        \STATE $\mathbf{R} = \textrm{softmax}(\gamma*\mathbf{W}/\mathbf{W}.\textrm{norm}(-1), -1)$
        \STATE $\mathbf{F}_\mathrm{IGT} = \textrm{einsum}("\textrm{np, npd}-> \textrm{nd}", \mathbf{R}, \mathbf{F}_\mathrm{t})$
        \STATE $\mathrm{Classifier_{IGT}} = \textrm{TextClassifierBuilder}(\mathbf{F}_\mathrm{IGT})$
        
        \STATE $\textcolor{gray}{\#\textrm{Inference}}.$
        \STATE $\mathrm{logits} = \mathrm{Classifier_{TGI}(f_v)} + \alpha * \mathrm{Classifier_{IGT}(f_v)}.$
        
        \STATE \textbf{return} logits
    \end{algorithmic}
\end{algorithm}

\subsection{B. Additional Experiments} \label{Additional Experiments}
\subsubsection{Complete Result On Each Dataset} We show a complete performance comparison of our method with others on 11 datasets, and the results are shown in Fig. \ref{LineChartPerformance}.

\subsubsection{Significance Analysis} \label{Significance Analysis}
To demonstrate that TIMO significantly outperforms the previous best method, GDA-CLIP, we conducted experiments with 100 random seeds. The random seed will affect the few-shot samples chosen as the support set. Initially, we formulate the null hypothesis, positing no significant difference between TIMO and GDA-CLIP. Subsequently, through experimentation, we obtained an average accuracy of 70.01\% for GDA-CLIP and 71.53\% for TIMO across 100 trials. Employing the Kruskal-Wallis test, we compute a p-value of 9.70e-34. We reject the null hypothesis at a significance threshold of 0.05, thus establishing that TIMO significantly outperforms GDA-CLIP.

\subsubsection{Model Fine-tuning} 
Following the paradigm of previous work \citep{Zhang_2022_Tip-Adapter, Zhu_2023_Not}, we use TaskRes \citep{Yu_2023_Task} to fine-tune our models. In detail, we keep the original TGI and IGT classifiers frozen and obtain a new classifier for the target task by tuning a set of zero Initialised parameters as a residual to the original one. The results after fine-tuning are shown in Tab. \ref{finetune}. We call the fine-tuned TIMO as TIMO-F.

From the table, it can be observed that although fine-tuning the model yields positive effects in some cases, overall, the performance of the fine-tuned models tends to decrease. Therefore, future research needs to explore new training methods to unleash the full potential of the Language-Image model.

\begin{figure}[!h]
    \centering
    \includegraphics[width=1\columnwidth]{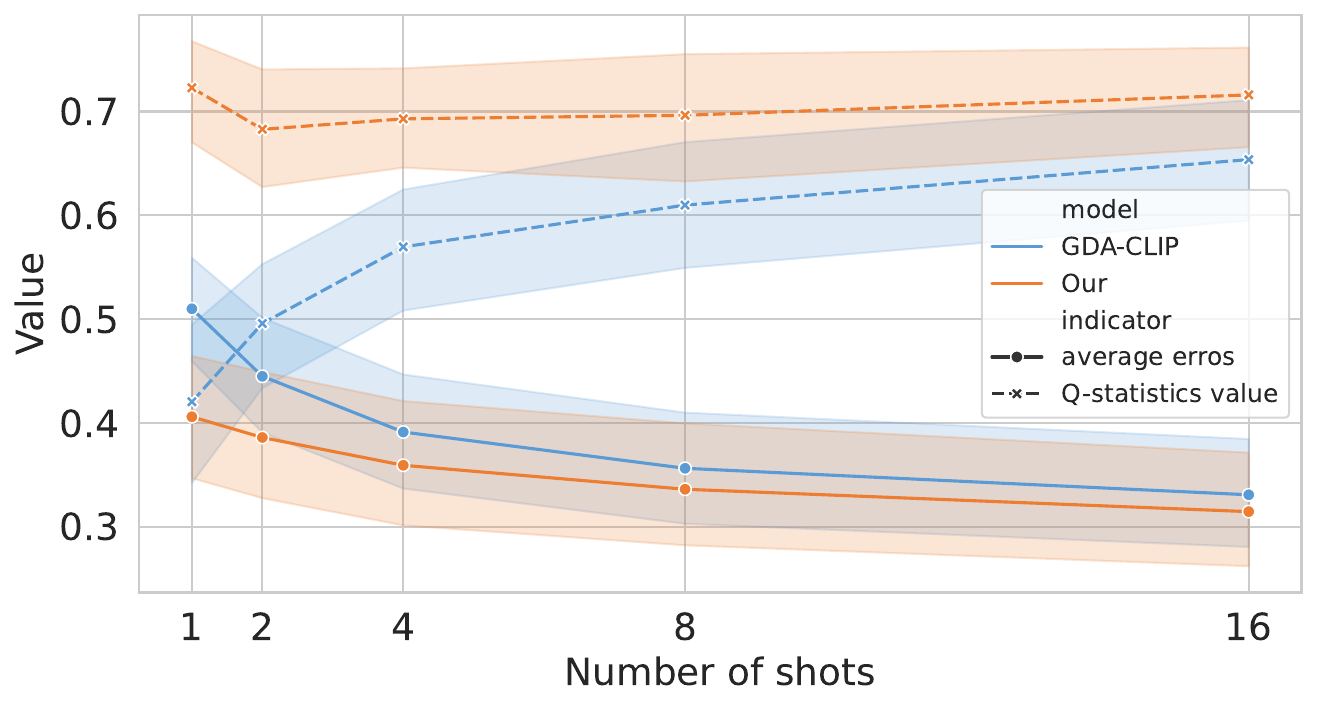}
    \caption{Comparison of generalization error and diversity between the GDA-CLIP and TIMO-S.} 
    \label{BoxPlotDiversity}
\end{figure}

\subsubsection{Ensemble Analysis}
From the perspective of ensemble learning, \citet{Krogh_1994_Neural} highlighted two conditions for effective performance: (1) diversity among individuals and (2) absence of flaws in individuals. We use Q-statistic to quantify diversity, with lower values indicating higher diversity, as shown in Fig. \ref{BoxPlotDiversity}. In this visualisation, lower Q-statistic values indicate higher diversity. The line graph suggests that our method's reduced error rate results from sacrificing diversity among individual classifiers. Future work could explore this trade-off further to identify an optimal balance point.

\subsubsection{Text Engagement in TGI} \label{Text engagement in TGI}
In constructing the TGI classifier, we integrate textual prompts with the image-based classifier, but we consider how text engagement may influence the final performance. To investigate this aspect, we conduct the following experiments: (1) Using all prompts to build TGI Classifier; (2) Using all prompts with weight (\emph{i.e.}, TIMO); (3) Using an adaptive number of prompts with weight (\emph{i.e.}, TIMO-S).

The experimental results are presented in Tab. \ref{TextEngagement}. Selectively using partial prompts for downstream tasks with weight could effectively enhance the final classification performance.

\subsubsection{Impact of Validation Size}
\citet{Lin_2023_Multimodality} and \citet{Silva-Rodriguez_2024_Closer} highlight that previous work overly relies on large validation sets, risking implicit knowledge disclosure and failing to demonstrate FSL's real-world relevance. Thus, we experiment to assess the validation set's impact on performance.

\begin{figure}[!h]
\centering
\includegraphics[width=1\columnwidth]{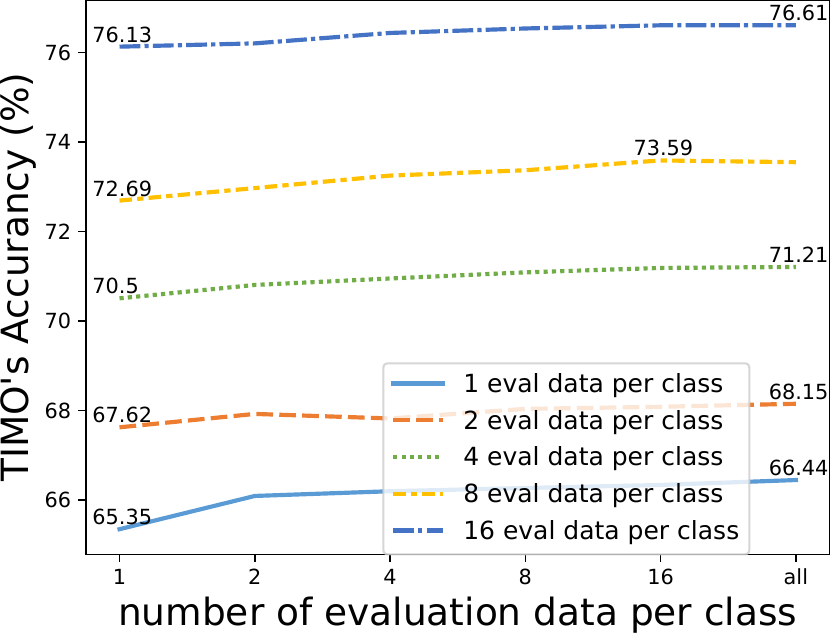}
\caption{Ablation of how the size of validation set affects the performance.} \label{EvalAblation}
\end{figure}

In Fig. \ref{EvalAblation}, we illustrate the average performance of TIMO across various evaluation sets and shot settings, highlighting both the best and worst performances. Overall, in scenarios with a larger validation set, TIMO achieves higher accuracy. However, beyond a certain number of shots (\textit{e.g.}, 16-shot), the impact of validation set size becomes minimal.

\subsection{C. Additional Experiment Details} \label{Additional Experiment Details}
\subsubsection{Employed Datasets and Associated Licences} \label{Employed Datasets and Associated Licences} We use the following datasets in this paper with their corresponding licenses. 
Caltech101 \citep{Fei-Fei_2004_Learning} is under CC BY 4.0 License. OxfordPets \citep{Parkhi_2012_Cats} is under CC BY-SA 4.0 License. 
EuroSAT \citep{Helber_2019_EuroSAT}, ImageNet-V2 \citep{Recht_2019_ImageNet}, ImageNet-Sketch \citep{Wang_2019_Learning}, ImageNet-A \citep{Hendrycks_2021_Natural} and ImageNet-R \citep{Hendrycks_2021_Many} are under MIT License. 
We could not find the licences of ImageNet \citep{Deng_2009_ImageNet}, FGVCAircraft \citep{Maji_2013_Fine-Grained}, StanfordCars \citep{Krause_2013_3D},  DTD \citep{Cimpoi_2014_Describing}, SUN397 \citep{Xiao_2010_SUN}, Flowers102 \citep{Nilsback_2008_Automated}, Food101 \citep{Bossard_2014_Food-101} and UCF101 \citep{Soomro_2012_UCF101} but they are publically free to researchers for non-commercial use.

\subsubsection{Experiment Setting} 
We construct our FSL model following the setup outlined in previous work \citep{Zhang_2022_Tip-Adapter}, where the validation set is commonly used for hyper-parameter search. However, CLAP \citep{Silva-Rodriguez_2023_Closer} suggests avoiding using a validation set in the traditional training process is possible. In our comparative analysis, we cite the results presented in Tab. \ref{PerformanceTable} of its original work, specifically CLAP's performance with a few-shot validation set. It should be noted that we also attempted to replicate the results using the exact same training and validation sets as in our method, but found that our replicated results were significantly lower than those reported in their paper. Therefore, we ultimately chose to compare our method against the performance results provided in their original paper.

Additionally, the default range of 3 introduced hyper-parameters in our work is $\alpha \in \{10^{-4},10^{-3},\ldots,10^{4}\}$, $\beta \in \{1,2,\ldots,2P\} $ and $\gamma \in \{5,10,15,\ldots,100\}$. Where $P$ is the number of prompts per category. If $\beta$ exceeds $P$, it implies repeated contributions of parts of $\mathbf{F}_\mathrm{t}$ to $\mathbf{F}_\mathrm{TGI}$ construction, achieved by pre-emptively repeating $\mathbf{F}_\mathrm{t}$ along the prompt dimension.

\subsubsection{Detailed Prompt Templates} \label{Detailed Prompt Templates}
We follow \citet{Zhu_2023_Not} to use both CuPL \citep{prattWhatDoesPlatypus2023} and Template-based Prompt \citep{Radford_2021_Learning} in our main experiments, as shown in Tab. \ref{Template-based Prompt} and Tab. \ref{CuPL Prompt}. Furthermore, we introduce two other prompt design methods, DCLIP \citep{Menon_2022_Visual} and WaffleCLIP \citep{Roth_2023_Waffling}, to research the effect of different prompt templates on our IGT classifier. Their design methods are described in detail in Tab. \ref{DCLIP} and Tab. \ref{WaffleCLIP}.

\subsection{D. Additional Case Studies} \label{Additional Visualization}
Although Fig. \ref{Illustration} briefly shows examples of anomalous matches and varying quality of generated prompts, we provide additional visualizations here.

\begin{figure}[!h]
    \centering
    \includegraphics[width=1\columnwidth]{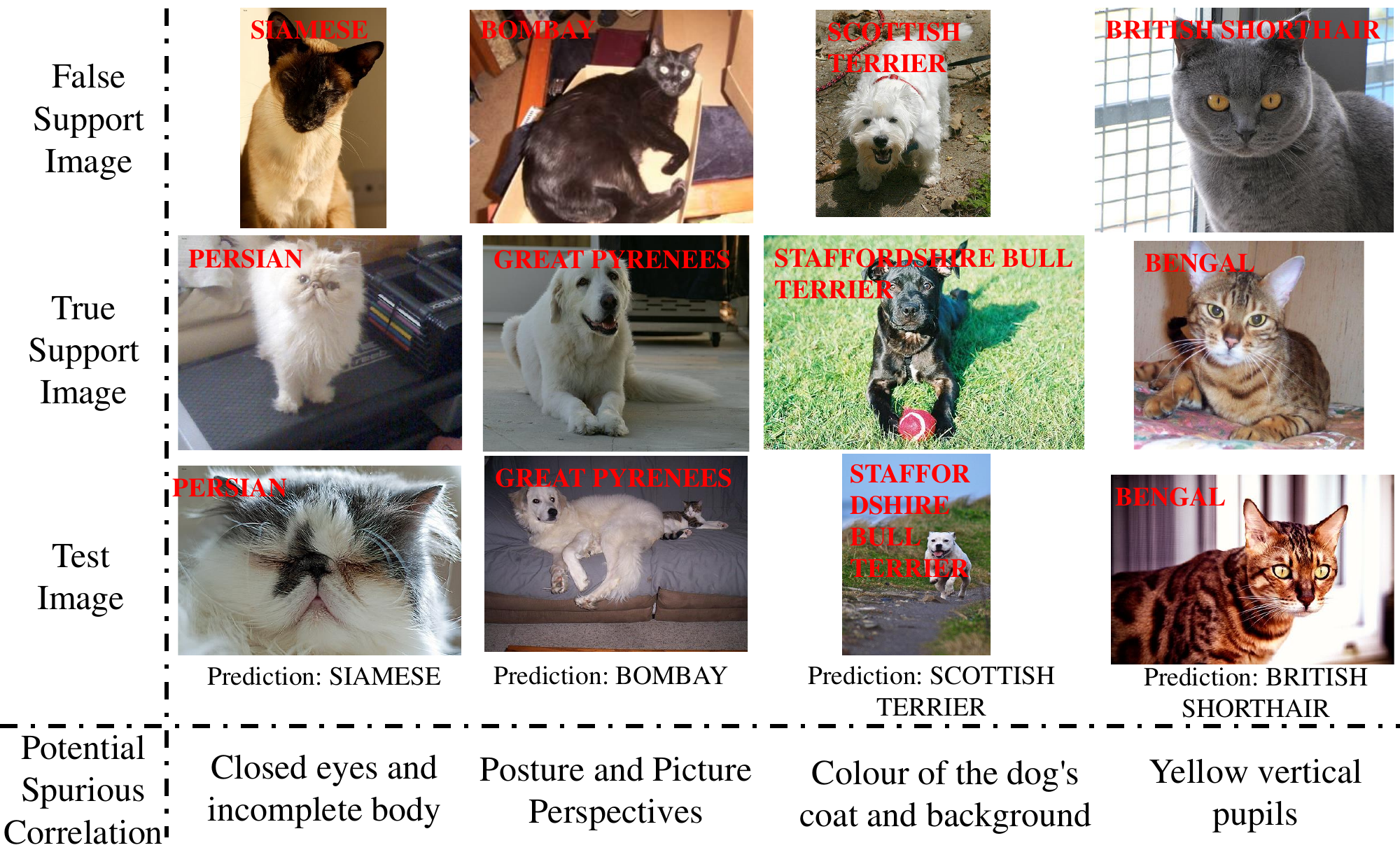}
    \caption{Anomalous matches with potential spurious correlation.} \label{FewShotProblem}
\end{figure}

1) \textit{anomalous matches} often result from spurious correlations like local or background similarities, as shown in Fig. \ref{FewShotProblem}.

2) \textit{The varying quality of generated prompts} is characterized by nonsensical or low-information sentences from large language models. (e.g., "There is no one definitive answer to this question" or "The Caltech101 dataset contains images of various animals, including this keeshond.")

\subsection{E. Additional Discussion} \label{Additional Discussion}
We proposed a sophisticated cross-modal mutual guidance mechanism that fully leverages complementary information between modalities to achieve enhanced performance. In-depth analysis suggests that our training-free mutual guidance idea may also be applicable to other fields.

Composed Image Retrieval (CIR) allows users to search for images by providing a combination of a reference image and additional textual modifications. However, current CIR methods are early in leveraging cross-modal interactions. For example, PromptCLIP \citep{Wu_2023_Few-Shot} uses image features as visual prompts for fine-tuning but overlooks CLIP's alignment, and zero-shot capabilities, and lacks in-depth cross-modal exploration. Besides, As model sizes grow, training becomes more challenging. Using a training-free strategy like TIMO in CIR allows adaptation to resource-constrained scenarios.

Black-box models fine-tuning aims to enhance performance in downstream tasks when both the model architecture and checkpoint are unknown, where CraFT \citep{Wang_2024_Connecting} enhances model inputs and outputs jointly. However, we find cross-modal guidance is also crucial in black-box scenarios. Our IGT module effectively merges prompts regardless of model architecture, indicating potential expansion directions.

\begin{table*}[!htb]
    \centering
    \setlength{\tabcolsep}{7.5mm} {
        \begin{tabular}{*{7}{c}}
           \toprule
           methods & 1-shot & 2-shot & 4-shot & 8-shot & 16-shot & Avg. \\
           \midrule
           Prompt       & 65.70 & 67.78 & 70.56 & 73.20 & 76.17 & 70.68 \\
           Prompt+w     & 66.43 & 68.15 & 71.20 & 73.55 & 76.61 & 71.19 \\
           \rowcolor{gray!40}  
           Prompt+w+s   & \textbf{66.82} & \textbf{68.65} & \textbf{71.61} & \textbf{74.10} & \textbf{76.81} & \textbf{71.60} \\
           \bottomrule
        \end{tabular}
    }
    \caption{Ablation study (\%) on how prompts are involved in the construction of TGI. Here, "Prompt" indicates using all prompts directly, "+w" signifies assigning different weights to each prompt, and "+s" represents using a validation set to adaptive search the number of prompts to use.} \label{TextEngagement}
\end{table*}

\begin{table*}[!htb]
    \centering
    \setlength{\tabcolsep}{5.8mm} {
        \begin{tabular}{*{8}{c}}
           \toprule
           \multirow{2}*{Method} & \multirow{2}*{Training} & \multicolumn{5}{c}{Number of shots} & \multirow{2}*{Avg.} \\
           \cmidrule(lr){3-7} 
           & & 1 & 2 & 4 & 8 & 16 \\
           \midrule
           TIMO (Ours)         &\XSolidBrush & \textbf{66.44} & 68.15 & \textbf{71.21} & 73.54 & \textbf{76.61} & \textbf{71.19} \\
           TIMO-F (Ours)       &\CheckmarkBold & 65.99 & \textbf{68.54} & 70.72 & \textbf{74.01} & 76.27 & 71.11 \\
           \bottomrule
        \end{tabular}
    }
    \caption{Average classification accuracy (\%) on 11 datasets of fine-tuned TIMO and TIMO-S.} 
    \label{finetune}
\end{table*}

\begin{table*}[!htb]
    \centering
    \setlength{\tabcolsep}{21.6mm} {
        \begin{tabular}{*{2}{c}}
           \toprule
           Dataset & DCLIP \\
           \midrule
           ALL  & A photo of a \{CLASS\}, which is/has/etc \{Descriptors\} \\
           \bottomrule
        \end{tabular}
        }
    \centering
    \caption{GPT-3 commands used in DCLIP \citep{Menon_2022_Visual} for each dataset.} \label{DCLIP}
\end{table*}

\begin{table*}[!htb]    
    \centering
    \setlength{\tabcolsep}{14mm} {
        \begin{tabular}{*{2}{c}}
           \toprule
           Dataset & WaffleCLIP \\
           \midrule
           \multirow{2}*{ALL}  
           & "A photo of a \{Concept\}, a \{CLASS\}, which is/has/etc \{Random Sequence\}"  \\
           & "A photo of a \{Concept\}, a \{CLASS\}, which is/has/etc \{Descriptor\}"  \\
           \bottomrule
        \end{tabular}
    }
    \caption{GPT-3 commands used in WaffleCLIP \citep{Roth_2023_Waffling} for each dataset.} \label{WaffleCLIP}
\end{table*}

\begin{table*}[!htb]
    \centering
    \setlength{\tabcolsep}{23.9mm} {
        \begin{tabular}{*{2}{c}}
           \toprule
           Dataset & Template Prompt \\
           \midrule
           \multirow{7}*{ImageNet}  
           & “itap of a \{CLASS\}.”  \\
           & “a bad photo of the \{CLASS\}.” \\
           & “a origami \{CLASS\}.”  \\
           & “a photo of the large \{CLASS\}.”  \\
           & “a \{CLASS\} in a video game.”  \\
           & “art of the \{CLASS\}.”  \\
           & “a photo of the small \{CLASS\}.”  \\
           \midrule
           \multirow{1}*{Caltech101}  
           & “a photo of a \{CLASS\}.”  \\
           \midrule
           \multirow{1}*{DTD}  
           & “\{CLASS\} texture.” \\
           \midrule
           \multirow{1}*{EuroSAT}  
           & “a centered satellite photo of \{CLASS\}.”  \\
           \midrule
           \multirow{1}*{FGVCAircraft}  
           & “a photo of a \{CLASS\}, a type of aircraft.”  \\
           \midrule
           \multirow{1}*{Flowers102}  
           & “a photo of a \{CLASS\}, a type of flower.”  \\
           \midrule
           \multirow{1}*{Food101}  
           & “a photo of \{CLASS\}, a type of food.”  \\
           \midrule
           \multirow{1}*{OxfordPets}  
           & “a photo of a \{CLASS\}, a type of pet.”  \\
           \midrule
           \multirow{1}*{StanfordCars}  
           & “a photo of a \{CLASS\}.”  \\
           \midrule
           \multirow{1}*{SUN397}  
           & “Describe what a \{CLASS\} looks like”  \\
           \midrule
           \multirow{1}*{UCF101}  
           & “a photo of a person doing \{CLASS\}.”  \\
           \bottomrule
        \end{tabular}
    }
    \caption{Template-based prompt for each dataset.} \label{Template-based Prompt}
\end{table*}

\begin{table*}[!htb]    
    \centering
    \setlength{\tabcolsep}{19mm} {
        \begin{tabular}{*{2}{c}}
           \toprule
           Dataset & GPT-3 Commands \\
           \midrule
           \multirow{5}*{ImageNet}  
           & “Describe what a \{CLASS\} looks like”  \\
           & “How can you identify \{CLASS\}?” \\
           & “What does \{CLASS\} look like?”  \\
           & “Describe an image from the internet of a \{CLASS\}”  \\
           & “A caption of an image of \{CLASS\}:”  \\
           \midrule
           \multirow{5}*{Caltech101}  
           & “Describe what a \{CLASS\} looks like”  \\
           & “How can you identify \{CLASS\}?” \\
           & “What does \{CLASS\} look like?”  \\
           & “Describe an image from the internet of a \{CLASS\}”  \\
           & “A caption of an image of \{CLASS\}:”  \\
           \midrule
           \multirow{6}*{DTD}  
           & “What does a \{CLASS\} material look like?”  \\
           & “What does a \{CLASS\} surface look like?”  \\
           & “What does a \{CLASS\} texture look like?”  \\
           & “What does a \{CLASS\} object look like?”  \\
           & “What does a \{CLASS\} thing look like?”  \\
           & “What does a \{CLASS\} pattern look like?”  \\
           \midrule
           \multirow{3}*{EuroSAT}  
           & “Describe an aerial satellite view of \{CLASS\}”  \\
           & “How does a satellite photo of a \{CLASS\} look like”  \\
           & “Visually describe a satellite view of a \{CLASS\}”  \\
           \midrule
           FGVCAircraft & “Describe a \{CLASS\} aircraft”  \\
           \midrule
           \multirow{4}*{Flowers102}  
           & “What does a \{CLASS\} flower look like”  \\
           & “Describe the appearance of a \{CLASS\}”  \\
           & “A caption of an image of \{CLASS\}”  \\
           & “Visually describe a \{CLASS\}, a type of flower”  \\
           \midrule
           \multirow{3}*{Food101}  
           & “Describe what a \{CLASS\} looks like”  \\
           & “Visually describe a \{CLASS\}”  \\
           & “How can you tell the food in the photo is a \{CLASS\}?”  \\
           \midrule
           \multirow{2}*{OxfordPets}  
           & “Describe what a \{CLASS\} pet looks like”  \\
           & “Visually describe a \{CLASS\}, a type of pet”  \\
           \midrule
           \multirow{9}*{StanfordCars}  
           & “How can you identify a \{CLASS\}”  \\
           & “Description of a \{CLASS\}, a type of car”  \\
           & “A caption of a photo of a \{CLASS\}:”  \\
           & “What are the primary characteristics of a \{CLASS\}?”  \\
           & “Description of the exterior of a \{CLASS\}”  \\
           & “What are the characteristics of a \{CLASS\}, a car?”  \\
           & “Describe an image from the internet of a \{CLASS\}”  \\
           & “What does a \{CLASS\} look like?”  \\
           & “Describe what a \{CLASS\}, a type of car, looks like”  \\
           \multirow{3}*{SUN397}  
           & “Describe what a \{CLASS\} looks like”  \\
           & “How can you identify a \{CLASS\}?”  \\
           & Describe a photo of a \{CLASS\}”  \\
           \midrule
           \multirow{3}*{UCF101}  
           & “What does a person doing \{CLASS\} look like”  \\
           & “Describe the process of \{CLASS\}”  \\
           & “How does a person \{CLASS\}”  \\
           \bottomrule
        \end{tabular}
    }
    \caption{GPT-3 commands used in CuPL for each dataset.} \label{CuPL Prompt}
\end{table*}